\documentclass[5p,times]{elsarticle}
\usepackage{amssymb, amsmath}
\usepackage{bbold}
\usepackage[mathscr]{euscript}
\usepackage{graphicx}
\usepackage[font=small,skip=5pt]{caption}
\usepackage{multirow}
\usepackage{subcaption}
\usepackage{adjustbox}
\usepackage{lipsum}
\usepackage{csquotes}
\nocite{*}
\usepackage{changes}
\newcommand{\w}{\text{\bbfamily w}}
\newcommand{\e}{\text{\bbfamily e}}
\newcommand{\bhyper}{\text{\bbfamily b}}

\begin{document}

\begin{frontmatter}
	
	\title{Fuzzy Least Squares Twin Support Vector Machines}
	
	\author[address1]{Javad Salimi Sartakhti\corref{mycorrespondingauthor}}
	\cortext[mycorrespondingauthor]{Corresponding author}
	
	\author[address2]{Homayun Afrabandpey}
	
	\author[address3]{Nasser Ghadiri}
	
	\address[address1]{Department of Electrical and Computer Engineering, University of Kashan, Kashan, Iran}
	\address[address2]{School of Science, Department of Computer Science, Aalto University, Finland}
	\address[address3]{Department of Electrical and Computer Engineering, Isfahan University of Technology (IUT), Isfahan, Iran}
		
	\begin{abstract}
		Least Squares Twin Support Vector Machine (LST-SVM) has been shown to be an efficient and fast algorithm for binary classification. It combines the operating principles of Least Squares SVM (LS-SVM) and Twin SVM (T-SVM); it constructs two non-parallel hyperplanes (as in T-SVM) by solving two systems of linear equations (as in LS-SVM). Despite its efficiency, LST-SVM is still unable to cope with two features of real-world problems. First, in many real-world applications, labels of samples are not deterministic; they come naturally with their associated membership degrees. Second, samples in real-world applications may not be equally important and their importance degrees affect the classification. In this paper, we propose Fuzzy LST-SVM (FLST-SVM) to deal with these two characteristics of real-world data. Two models are introduced for FLST-SVM: the first model builds up crisp hyperplanes using training samples and their corresponding membership degrees. The second model, on the other hand, constructs fuzzy hyperplanes using training samples and their membership degrees. Numerical evaluation of the proposed method with synthetic and real datasets demonstrate significant improvement in the classification accuracy of FLST-SVM when compared to well-known existing versions of SVM.
	\end{abstract}
	
	
\end{frontmatter}


\section{Introduction}

Support Vector Machine (SVM) is a classification technique based on the idea of Structural Risk Minimization (SRM) \cite{cortes1995support}. The algorithm has been used in many applications such as text classification \cite{joachims1998text, haddoud2016combining}, image classification \cite{zhang2012twin, pasolli2014svm}, and bioinformatics  \cite{sartakhti2012hepatitis, ubeyli2008support, gillani2014comparesvm}. The central ides of SVM is to find the optimal separating hyperplane between the positive and negative samples. The optimal hyperplane is the one that provides maximum margin between the closest training samples and the hyperplane. Due to its popularity, several versions of SVM have been proposed, among which the most important ones are Least Squares SVM (LS-SVM) \cite{suykens1999least}, Proximal SVM (P-SVM) \cite{mangasarian2001proximal}, Generalized Eigenvalue Proximal SVM (GEP-SVM) \cite{mangas2006multi}, and Twin SVM (T-SVM) \cite{khemchandani2007twin}.

Least Squares Twin Support Vector Machine (LST-SVM) \cite{arun2009least} is a relatively new version of SVM, which combines the idea of LS-SVM and T-SVM. It determines two non-parallel hyperplanes by solving two systems of linear equations instead of non-linear ones. Although the algorithm provides high accuracies in some applications, both LST-SVM and its improved versions \cite{sartakhti2017simulated, tanveer2016robust} suffer from two main drawbacks: (I) they have the implicit assumption that the associated labels of samples are deterministic, while in many real-world applications, labels come naturally with uncertainties in the form of membership degrees. Example of such applications is ``spam filtering'' where it is difficult to deterministically assign each email to one of the two classes of ``spam'' or ``normal'' \cite{amayri2010study}. (II) In many classification tasks, data points might have different importances, while LST-SVM considers them to be equally important. This happens a lot in bioinformatics applications or other applications with unbiased class labels. A potential approach to cope with these challenges is to arm LST-SVM with fuzzy theory \cite{zadeh1975concept,dubois1980fuzzy}. Fuzzy theory provides useful tools when analyzing complex processes using standard quantitative methods or when the available information is interpreted uncertainly. A fuzzy function offers an efficient way of capturing the inexact nature of real-world problems by representing uncertainty in the data using fuzzy parameters.

In this paper we apply the fuzzy set theory to LST-SVM algorithm and propose a novel formulation called Fuzzy LST-SVM (FLST-SVM). The two key features of FLST-SVM are (I) assigning fuzzy membership values to data points based on their importance degrees, and (II) the parameters of the FLST-SVM model, e.g. the weight vector and the bias term, are fuzzified. Using these two features, we propose two models for FLST-SVM.
It should be noted that we are not the first to employ the fuzzy concept for improving SVM type algorithms. \cite{lin2002fuzzy,batuwita2010fsvm,wang2014interval,hao2008fuzzy} are examples of fuzzy formulation for SVM. Recently, Han and Coa proposed a fuzzy extension of LST-SVM \cite{han2017fuzzy} called Fuzzy Chance-Constrained LST-SVM (FCC-LSTSVM). However, there are several main differences between our method and their approach. First, FCC-LSTSVM only proposed a model with fuzzy membership degrees for samples, but the hyperplanes are crisp. In our formulation, all elements of the LST-SVM model are fuzzified. Second, they applied their model to LST-SVM with chance-constrained programming, which is a particular case of general LST-SVM. We, on the other hand, propose a fuzzy model for the general LST-SVM algorithm. Finally, our method significantly outperforms the FCC-LSTSVM (we did not evaluate the performance of FCC-LSTSVM in our experiments since we did not have access to the source codes, however results on some of UCI data sets are available in Table 3 of \cite{han2017fuzzy} and Table III of this paper).

The rest of this paper is organized as follows. A brief review of basic concepts including the SVM, T-SVM, and LST-SVM is presented in Section \ref{SecII}. Two fuzzy models of LST-SVM are introduced in Section \ref{SecIII}. In Section \ref{SecIV} we evaluate the proposed models, and finally, Section \ref{SecV} concludes the paper.

\section{Background}\label{SecII}

This section reviews some of the related versions of SVM, namely the standard SVM, T-SVM, and LST-SVM. In the rest of the paper, bold-face lower-case letters refer to vectors (e.g. $\pmb{a}$), bold-face capital letters refer to matrices (e.g. $\pmb{A}$), $||\pmb{A}||$ denotes the determinant of the matrix $\pmb{A}$, the transpose of a matrix (and/or vector) $\pmb{A}$ is shown with $\pmb{A}^{T}$, and finally a vector of zeros is shown with $\pmb{0}$.

\subsection{Support Vector Machine}

\begin{figure}[ht!]
	\centering
	\begin{subfigure}[b]{0.45\columnwidth}
		\includegraphics[width=\columnwidth]{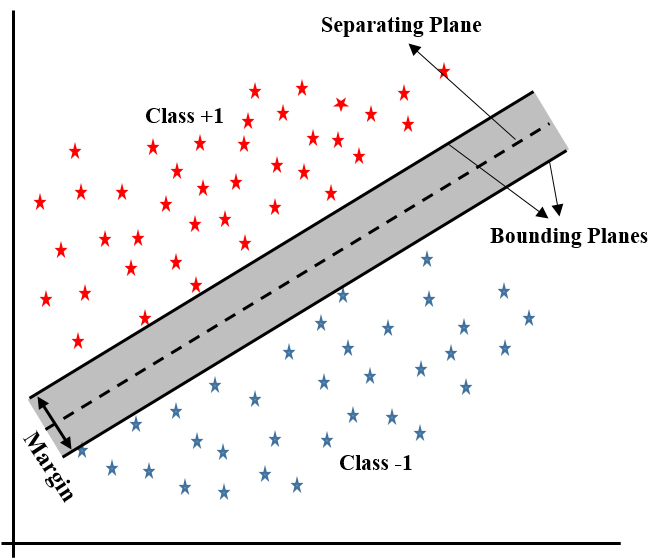}
		\caption{}
		\label{SVM}
	\end{subfigure}%
	\hspace{0.03\textwidth}%
	\begin{subfigure}[b]{0.45\columnwidth}
		\includegraphics[width=\columnwidth]{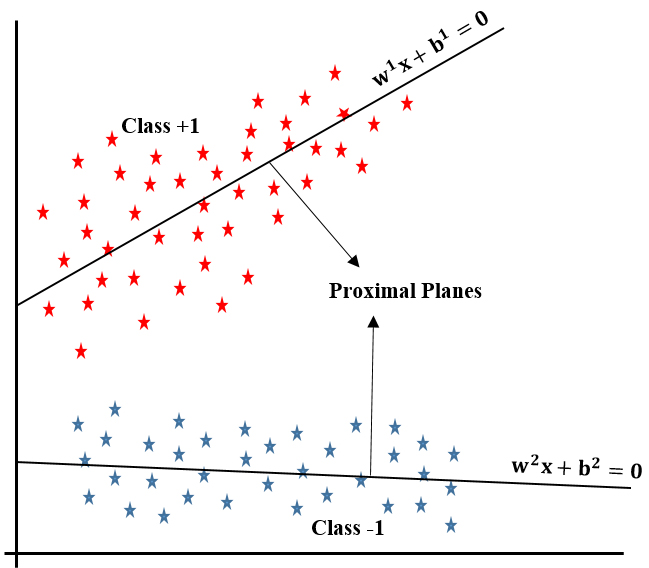}
		\caption{}
		\label{TSVM}
	\end{subfigure}%
	\caption{Geometric interpretation of (a) SVM and (b) T-SVM}
	\label{Fig4}
\end{figure}

The main idea in SVM is to minimize the classification error while preserving the maximum possible margin between classes. Suppose a binary classification task with a set of training samples $\{\pmb{x}_{s}\}_{s = 1}^{n}\in\mathbb{R}^{d}$ with their corresponding labels $y_{s}\in\lbrace-1,+1\rbrace$. SVM aims at finding a hyperplane with equation $\pmb{w}^{T}.\pmb{x}_{s}+b=0$ under the following constraints:
\begin{flalign*}
&y_{s}(\pmb{w}^{T}.\pmb{x}_{s}+b) \geq1,\qquad \forall s&
\end{flalign*}
where the vector $\pmb{w} \in \mathbb{R}^d$ and the bias term $b \in \mathbb{R}$ are the parameters to be learned. The goal of the constraints is to ensure that samples are at a maximum distance (which is set to 1 in standard SVM) from the separating hyperplane. The parameters are obtained by solving the following constrained optimization problem known as \textit{primal} problem.
\begin{flalign}\label{eq1}
&\mbox{Minimize}\quad f(x)=\frac{\Vert \pmb{w}\Vert^{2}}{2}& \\\nonumber
&\mbox{subject to}\quad y_{s}(\pmb{w}^{T}.\pmb{x}_{s}+b)-1\geq 0&
\end{flalign}
In this equation, $\Vert\pmb{w}\Vert = \sqrt{w_{1}^{2} + \cdots + w_{d}^{2}}$ denotes the norm-2 of the $\pmb{w}$ vector. To reduce the complexity, typically the \textit{dual} form, a widely adopted trick in convex optimization, of equation \ref{eq1} is optimized. The geometric interpretation of SVM is depicted in Figure \ref{SVM} for a toy example. The bounding planes are hypothetical planes showing the maximum margin in each side of the separating plane.
\subsection{Twin Support Vector Machine}\label{SecII.2}
SVM finds a single hyperplane for classifying samples. Jayadeva et al. \cite{khemchandani2007twin} proposed T-SVM with the idea of finding two hyperplanes in which samples are assigned to a class according to their distance from the hyperplanes. Equations of the two hyperplanes are as follows:
\begin{flalign*}
&\pmb{w}_{1}^{T}\pmb{x}_{s}+b_{1}=0& \\\nonumber
&\pmb{w}_{2}^{T}\pmb{x}_{s}+b_{2}=0&
\end{flalign*}
where $\pmb{w}_{i}$ and $b_{i}$ denote the parameters of the $i^{th}$ hyperplane. Hyperplanes are non-parallel, and each of them is closest to the samples of its own class and farthest from the samples of the opposite class. The concept is geometrically depicted in Figure \ref{TSVM} for a toy example.

Assume a binary classification task with classes $+1$ and $-1$, and $\pmb{A} \in \mathbb{R}^{n_{1}\times d}$ and $\pmb{B} \in \mathbb{R}^{n_{2}\times d}$ indicate matrices of samples belonging to classes $+1$ and $-1$, respectively. Each row of these matrices denotes one sample of their corresponding class. The two hyperplanes of T-SVM obtained by solving equations \ref{eq2} and \ref{eq3}.
\begin{flalign}\label{eq2}
&\mbox{min} \quad \frac{1}{2}(\pmb{A}\pmb{w}_{1}+\pmb{e}_{1}b_{1})^{T}(\pmb{A}\pmb{w}_{1}+\pmb{e}_{1}b_{1})+p_{1}\pmb{e}_{2}^T\pmb{\xi}& \\\nonumber
&\mbox{s.t.} \quad -(\pmb{B}\pmb{w}_{1}+\pmb{e}_{2}b_{1})+\pmb{\xi} \geq \pmb{e}_{2}, \; \pmb{\xi} \geq \pmb{0}&
\end{flalign}
\begin{flalign}\label{eq3}
&\mbox{min} \quad \frac{1}{2}(\pmb{B}\pmb{w}_{2}+\pmb{e}_{2}b_{2})^{T}(\pmb{B}\pmb{w}_{2}+\pmb{e}_{2}b_{2})+p_{2}\pmb{e}^T_{1}\pmb{\xi}& \\\nonumber
&\mbox{s.t.} \quad \pmb{A}\pmb{w}_{2}+\pmb{e}_{1}b_{2}+\pmb{\xi} \geq \pmb{e}_{1}, \; \pmb{\xi} \geq \pmb{0}.&
\end{flalign}
In these equations, $\pmb{\xi}$ represents the vector of slack variables of size $n$, and $\pmb{\xi} \geq \pmb{0}$ means each component of this vector is non-negative. If the training samples are not linearly separable, the standard approach is to let the decision margins make a few mistakes (i.e. some points are inside or in the wrong side of the margin). Each non-zero element of the slack variables vector determines a cost for the misclassified sample which is proportional to the distance between the sample and the true decision margin. In the above equations, $\pmb{e}_{i} \; (i \in \{1,2\})$ is a column vector of ones with appropriate length, and $p_{1}$ and $p_{2}$ are penalty parameters.

\subsection{Least Squares Twin Support Vector Machine}\label{SecII.3}
\noindent
LST-SVM is a binary classifier, which combines the idea of LS-SVM and T-SVM. It converts the inequality constraints in T-SVM to equality constraints and solves two linear equations systems rather than two Quadratic Programming Problems (QPPs). Experiments have shown that LST-SVM can considerably reduce the training time, while providing competitive classification accuracy \cite{gao20111}. The time complexity of the standard SVM is of order $n^{3}$, where $n$ is the number of constraints (which is equal to the number of samples, i.e. one constraint per sample). Theoretically, when the number of samples in the two classes of a binary classification task are equal, LST-SVM is four times faster than standard SVM.

LSTSVM finds the separating hyperplanes by optimizing functions \ref{eq4} and \ref{eq5}, which are linearly solvable.
\begin{flalign}\label{eq4}
&\mbox{min} \quad \frac{1}{2}(\pmb{A}\pmb{w}_{1}+\pmb{e}b_{1})^{T}(\pmb{A}\pmb{w}_{1}+\pmb{e}b_{1})+ \frac{p_{1}}{2}\pmb{\xi}^{T}\pmb{\xi}& \\\nonumber
&\mbox{s.t.} \quad -(\pmb{B}\pmb{w}_{1}+\pmb{e}b_{1})+\pmb{\xi}=\pmb{e}&
\end{flalign}
\begin{flalign}\label{eq5}
&\mbox{min} \quad \frac{1}{2}(\pmb{B}\pmb{w}_{2}+\pmb{e}b_{2})^{T}(\pmb{B}\pmb{w}_{2}+\pmb{e}b_{2})+ \frac{p_{2}}{2}\pmb{\xi}^{T}\pmb{\xi}& \\\nonumber
&\mbox{s.t.} \quad (\pmb{A}\pmb{w}_{2}+\pmb{e}b_{2})+\pmb{\xi}=\pmb{e}&
\end{flalign}
Solving the above functions gives us parameters of the hyperplanes, i.e. $\pmb{w}$ and $b$, as follows:
\begin{flalign}\label{eq6}
&\begin{bmatrix}
	\pmb{w}_{1} \\
	b_{1}
\end{bmatrix}
=-(\pmb{F}^{T}\pmb{F} + \frac{1}{p_{1}} \pmb{E}^{T}\pmb{E})^{-1}\pmb{F}^{T}\pmb{e}&
\end{flalign}
\begin{flalign}\label{eq7}
&\begin{bmatrix}
	\pmb{w}_{2} \\
	b_{2}
\end{bmatrix}
=(\pmb{E}^{T}\pmb{E} + \frac{1}{p_{2}} \pmb{F}^{T}\pmb{F})^{-1}\pmb{E}^{T}\pmb{e}&
\end{flalign}
where $\pmb{E}=\begin{bmatrix}
\pmb{A} & \pmb{e}
\end{bmatrix}$ and $\pmb{F}= \begin{bmatrix}
\pmb{B} & \pmb{e}
\end{bmatrix}$ and $\pmb{A}$, $\pmb{B}$, $\pmb{e}$ and $\pmb{\xi}$ are already introduced in Section \ref{SecII.2}.

\section{Fuzzy Least Squares Twin Support Vector Machine}\label{SecIII}
\noindent
In many real-world applications, samples in the training data do not strictly belong to a single class. Furthermore, in some applications it is desirable to have different importance degrees for training samples, e.g. in recommender systems newer products should have higher importance degrees than older ones. Given the uncertainty of assigning such importance values, the fuzzy sets provide an elegant way to cope with this problem. A fuzzy membership degree $\mu_{s}$ can be defined for each sample $s$ in the training data. A membership degree is a number between $0$ and $1$ which determines to what extent a sample belongs to a class. Therefore, a training sample with membership degree of $\mu_{s}$ belongs to class $+1$ by $\mu_{s}$ and belongs to class $-1$ by $(1-\mu_{s})$.

Fuzzy SVM is first introduced in \cite{hao2008fuzzy}, where the author proposed two models, $M_{1}$ and $M_{2}$, for applying fuzzy sets in SVM. In the first model, $M_{1}$, a crisp hyperplane was learned using samples with fuzzy membership degrees. In the second model, $M_{2}$, a fuzzy hyperplane was obtained to discriminate classes. In the following, we apply the idea of \cite{hao2008fuzzy} to LST-SVM which has more parameters than SVM, that makes the modeling more complicated and the formulations need more careful attention.

\subsection{Fuzzy LST-SVM: Model $M_{1}$}

In this model, a fuzzy membership function is defined to assign fuzzy memberships degrees to samples such that noises and outliers acquire smaller values. Our goal is to construct two crisp hyperplanes to distinguish target classes. For this purpose, equations \ref{eq4} and \ref{eq5} are modified as follows:
\begin{flalign}\label{eq8}
&\mbox{min} \; J_{1}=\frac{1}{2}(\pmb{A}\pmb{w}_{1}+\pmb{e}b_{1})^{T}(\pmb{A}\pmb{w}_{1}+\pmb{e}b_{1})+\frac{p_{1}}{2}\pmb{\mu}_{1}^{T} \pmb{\xi}^{2}& \\\nonumber
&\mbox{s.t.} \; -(\pmb{B}\pmb{w}_{1}+\pmb{e}b_{1})+\pmb{\xi}=\pmb{e}&
\end{flalign}
\begin{flalign}\label{eq9}
&\mbox{min} \; J_{2}=\frac{1}{2}(\pmb{B}\pmb{w}_{2}+\pmb{e}b_{2})^{T}(\pmb{B}\pmb{w}_{2}+\pmb{e}b_{2})+\frac{p_{2}}{2}\pmb{\mu}_{2}^{T} \pmb{\xi}^{2}& \\\nonumber
&\mbox{s.t.} \; (\pmb{A}\pmb{w}_{2}+\pmb{e}b_{2})+\pmb{\xi}=\pmb{e}&
\end{flalign}
to represent the positive ($+1$) and negative ($-1$) class, respectively. In these equations, $\pmb{\mu}_{i}; \; i \in \{1,2\}$ is the vector of membership values of the samples of each class (i.e. in the first equation, $\pmb{\mu}_{1}$ is in $\mathbb{R}^{n_{1} \times 1}$ where $n_{1}$ is the number of samples in $\pmb{A}$, while $\pmb{\mu}_{2}$ is in $\mathbb{R}^{n_{2} \times 1}$), $\pmb{\xi}^{2}$ is the vector of the second power of the elements of $\pmb{\xi}$. Therefore, $\frac{p_{i}}{2}\pmb{\mu}^{T} \pmb{\xi}^{2}$ determines the total amount of cost/penalty for each decision margin based on the membership degrees of its samples. By rearranging the equations of the constraints, we can easily conclude the equation of $\pmb{\xi}$. By substituting $\pmb{\xi}$ with its equivalence, obtained from the constraints, we have
\begin{flalign}\label{eq10}
&\mbox{min} \; J_{1}=\frac{1}{2}\Vert \pmb{A}\pmb{w}_{1}+\pmb{e}b_{1}\Vert ^{2}+\frac{p_{1}}{2}\parallel \pmb{\mu}_{1}\parallel \Vert \pmb{B}\pmb{w}_{1}+\pmb{e}b_{1}+\pmb{e}\Vert^{2}&
\end{flalign}
\begin{flalign}\label{eq11}
&\mbox{min} \; J_{2}=\frac{1}{2}\Vert \pmb{B}\pmb{w}_{2}+\pmb{e}b_{2}\Vert ^{2}+\frac{p_{2}}{2}\parallel \pmb{\mu}_{2} \parallel \Vert \pmb{A}\pmb{w}_{2}+\pmb{e}b_{2}+\pmb{e}\Vert^{2}& \end{flalign}
where the parameters are $\{\pmb{w}_{i}, b_{i}\}$ for each function $J_{i}$. Differentiating the functions with respect to their parameters gives us
\begin{flalign*}
&\frac{\partial J_{1}}{\partial \pmb{w}_{1}}=\pmb{A}^{T}(\pmb{A}\pmb{w}_{1}+\pmb{e}b_{1})+p_{1}\pmb{\mu}_{1}\pmb{B}^{T}(\pmb{B}\pmb{w}_{1}+\pmb{e}b_{1}+\pmb{e})=\pmb{0}&
\end{flalign*}
\begin{flalign*}
&\frac{\partial J_{1}}{\partial b_{1}}=\pmb{e}^{T}(\pmb{A}\pmb{w}_{1}+\pmb{e}b_{1})+p_{1}\pmb{\mu}_{1}\pmb{e}^{T}(\pmb{B}\pmb{w}_{1}+\pmb{e}b_{1}+\pmb{e})=0&
\end{flalign*}
\begin{flalign*}
&\frac{\partial J_{2}}{\partial \pmb{w}_{2}}=\pmb{B}^{T}(\pmb{B}\pmb{w}_{2}+\pmb{e}b_{2})+p_{2}\pmb{\mu}_{2}\pmb{A}^{T}(\pmb{A}\pmb{w}_{2}+\pmb{e}b_{2}+\pmb{e})=\pmb{0}& 
\end{flalign*}
\begin{flalign*}
&\frac{\partial J_{2}}{\partial b_{2}}=\pmb{e}^{T}(\pmb{B}\pmb{w}_{2}+\pmb{e}\pmb{b}_{2})+p_{2}\pmb{\mu}_{2}\pmb{e}^{T}(\pmb{A}\pmb{w}_{2}+\pmb{e}b_{2}+\pmb{e})=0.&
\end{flalign*}
Solving the above equations using matrix algebra gives us the hyperplanes  of the positive and negative class shown in equations \ref{eq12} and \ref{eq13}, respectively.
\begin{flalign}\label{eq12}
&\begin{bmatrix}
\pmb{w}_{1} \\
b_{1}
\end{bmatrix}
=\begin{bmatrix}
\pmb{\mu}_{1}\pmb{B}^{T}\pmb{B}+\frac{1}{p_{1}}\pmb{A}^{T}\pmb{A} & \pmb{\mu}_{1}\pmb{B}^{T}\pmb{e}+\frac{1}{p_{1}}\pmb{A}^{T}\pmb{e} \\
\pmb{\mu}_{1}\pmb{e}^{T}\pmb{B}+\frac{1}{p{1}}\pmb{e}^{T}\pmb{A} & \pmb{\mu}_{1} m_{2}+\frac{1}{p_{1}}m_{1}\pmb{e}
\end{bmatrix}^{-1} \begin{bmatrix}
-\pmb{B}^{T}\pmb{e}\\
-m_{2}
\end{bmatrix}&
\end{flalign}
\begin{flalign}\label{eq13}
&\begin{bmatrix}
\pmb{w}_{2} \\
b_{2}
\end{bmatrix}
=\begin{bmatrix}
\pmb{\mu}_{2}\pmb{A}^{T}\pmb{A}+\frac{1}{p_{2}}\pmb{B}^{T}\pmb{B} & \pmb{\mu}_{2}\pmb{A}^{T}\pmb{e}+\frac{1}{p_{2}}\pmb{B}^{T}\pmb{e} \\
\pmb{\mu}_{2}\pmb{e}^{T}\pmb{A}+\frac{1}{p{2}}\pmb{e}^{T}\pmb{B} & \pmb{\mu}_{2} m_{1}+\frac{1}{p_{2}}m_{2}\pmb{e}
\end{bmatrix}^{-1} \begin{bmatrix}
-\pmb{A}^{T}\pmb{e}\\
-m_{1}
\end{bmatrix}&
\end{flalign}
In these equations, $m_1$ and $m_2$ are the numbers of constraints (equivalently the number of samples) in the first and the second class, respectively. Once the values of the parameters $\{\pmb{w}_{i}, b_{i}\}_{i=1}^{2}$ are obtained, a new input sample is assigned to a class based on its distance from the hyperplane of the corresponding class.

\subsection{Fuzzy LST-SVM: Model $M_{2}$}

In this model, we construct fuzzy hyperplanes to discriminate the classes. All parameters of the model, even the components of $\pmb{w}$, are fuzzy variables. For the sake of computational simplicity, parameters are restricted to a class of \textit{triangular} symmetric membership function. A symmetric triangular fuzzy number $X$ is shown as $X = \prec o,r\succ$ where $o$ is center and $r$ is width of the corresponding membership function.

We define $\w_{i} = \{\left(\pmb{w}_{i}, \pmb{c}_{i}\right)\}$ and $\bhyper_{i} = \{(b_i,d_i)\}$ for each fuzzy hyperplane, where $\pmb{c}_{i} = \{c_{ij}\}_{i \in \{1,2\}, j=1, \cdots,d}$ is the vector of fuzzy degrees of each component of $\pmb{w}_i$, and $d_i$ is the fuzzy degree of the bias term. Then the equation of each fuzzy hyperplane can be written as:
\begin{flalign}\label{eq14}
&\w_{i}^{T}.\pmb{x}_s+\bhyper_{i}=\prec w_{i1},c_{i1}\succ .x_{s1}+\cdots+ \prec w_{id},c_{id}\succ .x_{sd}+\prec b_{i},d_{i}\succ =0&
\end{flalign}
where each $\prec .,.\succ$ is a symmetric triangular fuzzy number as defined above, and inner product of a scalar with a symmetric triangular fuzzy number results in another symmetric triangular fuzzy number with center and width multiplied by the scalar. To find the fuzzy hyperplane for class $+1$, we rewrite equation \ref{eq8} as:
\begin{flalign}\label{eq15}
&\mbox{min} \; J=\frac{1}{2}(\pmb{A}\w_{1}+\pmb{e}\bhyper_{1})^{T}(\pmb{A}\w_{1}+\pmb{e}\bhyper_{1})+ \frac{p_{1}}{2}\pmb{\mu}_{1}^{T} \pmb{\xi}^{2}+\tau(\frac{1}{2}\Vert \pmb{c}_{1} \Vert ^{2} +d_{1})& \\\nonumber
&\mbox{s.t.} \quad -(\pmb{B}\w_{1} +\pmb{e}\bhyper_{1})=\pmb{\e}-\pmb{\xi}&
\end{flalign}
where $\e$ is a fuzzy margin which is a vector of fuzzy numbers each with center one and width $O_{\w_{1}}$, and $\tau$ is a user-selected scalar control parameter which is set to $1$ in our implementation. In this equation, $\frac{1}{2}\Vert \pmb{c}_{1} \Vert ^{2}+d_{1}$ measures the vagueness of the model. More vagueness means that the decision boundary is more unclear around a given point while less vagueness leads to strict boundary. The fuzzy constraint in equation \ref{eq15} can be written in terms of two constraints, one for each side of the center, as follows.
\begin{flalign}\label{eq161}
&\mbox{min} \; J=\frac{1}{2}(\pmb{A}\w_{1}+\pmb{e}\bhyper_{1})^{T}(\pmb{A}\w_{1}+\pmb{e}\bhyper_{1})+ \frac{p_{1}}{2}\pmb{\mu}_{1}^{T} (\pmb{\xi}_{1}+\pmb{\xi}_{2})^{2} + \tau(\frac{1}{2}\Vert \pmb{c}_{1}\Vert ^{2}+d_{1})& \\\nonumber
&\mbox{s.t.} \quad (\pmb{B}\pmb{w}_{1}+\pmb{e}b_{1})+(\pmb{B}\pmb{c}_{1}+\pmb{e}d_{1})=\pmb{e}+O_{\w_{1}}-\pmb{\xi}_{1}& \\\nonumber
& \quad\quad\quad (\pmb{B}\pmb{w}_{1}+\pmb{e}b_{1})-(\pmb{B}\pmb{c}_{1}+\pmb{e}d_{1})=\pmb{e}-O_{\w_{1}}-\pmb{\xi}_{2}&
\end{flalign}
By substituting the slack variable with its equivalence, obtained from the constraints, as in equations \ref{eq10} and \ref{eq11}, the hyperplane of the first class is obtained by optimizing equation \ref{eq16} with respect to its parameters, i.e. $\pmb{w}_{1}, b_1, \pmb{c}_1,$ and $d_1$.
\begin{flalign}\label{eq16}
	\includegraphics[scale=0.55]{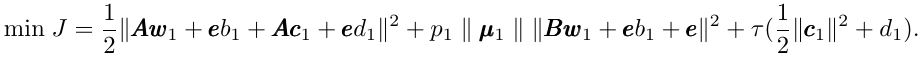}
\end{flalign}
Setting the derivation of function \ref{eq16} with respect to each parameter equal to zero, we will have:
\begin{flalign*}
&\frac{\partial J}{\partial \pmb{w}_{1}}=\pmb{A}^{T}(\pmb{A}\pmb{w}_{1}+\pmb{e}b_{1}+\pmb{A}\pmb{c}_{1}+\pmb{e}d_{1})+ p_{1}\pmb{\mu}_{1} \pmb{B}^{T}(\pmb{B} \pmb{w}_{1}+\pmb{e}b_{1}+\pmb{e})=\pmb{0}&
\end{flalign*}
\begin{flalign*}
&\frac{\partial J}{\partial b_{1}}=\pmb{e}^{T}(\pmb{A}\pmb{w}_{1}+\pmb{e}b_{1}+\pmb{A}\pmb{c}_{1}+ \pmb{e}d_{1})+p_{1}\pmb{\mu}_{1} \pmb{e}^{T}(\pmb{B} \pmb{w}_{1}+\pmb{e}b_{1}+\pmb{e})=0&
\end{flalign*}
\begin{flalign*}
&\frac{\partial J}{\partial \pmb{c}_{1}}=\pmb{A}^{T}(\pmb{A}\pmb{w}_{1}+\pmb{e}b_{1}+\pmb{A}\pmb{c}_{1}+\pmb{e}d_{1})+\tau\pmb{c}_{1}=\pmb{0}.&
\end{flalign*}
Finally, the parameters of the hyperplane is given by:
\begin{equation}
\resizebox{\hsize}{!}{

$ \left(
  \begin{array}{@{}*6r}
 \pmb{w}_{1} \\
 b_{1} \\
 \pmb{c}_{1} \\
 d_{1}\\
  \end{array}
\right)=
\left(
  \begin{array}{@{}*6r}
\frac{1}{p_{1}}\pmb{A}^{T}\pmb{A}+\pmb{\mu}_{1} \pmb{B}^{T}\pmb{B} & \frac{1}{p_{1}}\pmb{A}^{T}\pmb{e}+\pmb{\mu}_{1} \pmb{B}^{T}\pmb{e} & \frac{1}{p_{1}}\pmb{A}^{T}\pmb{A} & \frac{1}{p_{1}}\pmb{A}^{T}\pmb{e} \\
\frac{1}{p_{1}}\pmb{e}^{T}\pmb{A}+\pmb{\mu}_{1} \pmb{e}^{T}\pmb{B} & \frac{1}{p_{1}}m_{1}+\pmb{\mu}_{1} m_{2} & \frac{1}{p_{1}}\pmb{e}^{T}\pmb{A} & \frac{1}{p_{1}}m_{1} \\
\pmb{A}^{T}\pmb{A} & \pmb{A}^{T}\pmb{e} & \pmb{A}^{T}\pmb{A}+\pmb{e}\tau\pmb{e}^{T} & \pmb{A}^{T}\pmb{e} \\
\pmb{e}^{T}\pmb{A} & m_{1} & \pmb{e}^{T}\pmb{A} & m_{1}\\
  \end{array}
\right)^{-1} \left(
  \begin{array}{@{}*6r}
 \pmb{\mu}_{1} \pmb{B}^{T}\pmb{e} \\
 \pmb{\mu}_{1} \\
 0 \\
 \tau\\
  \end{array}
\right)
$
}
\end{equation}
Thus far, we have obtained the first fuzzy hyperplane. In a similar way the second hyperplane is given by:
\begin{equation}
\resizebox{\hsize}{!}{

$ \left(
  \begin{array}{@{}*6r}
 \pmb{w}_{2} \\
 b_{2} \\
 \mathbb{c}_{2} \\
 d_{2}\\
  \end{array}
\right)=
\left(
  \begin{array}{@{}*6r}
\pmb{\mu}_{2} \pmb{A}^{T}\pmb{A}+\frac{1}{p_{2}}\pmb{B}^{T}\pmb{B} & \pmb{\mu}_{2} \pmb{A}^{T}\pmb{e}+\frac{1}{p_{2}}\pmb{B}^{T}\pmb{e} & \frac{1}{p_{2}}\pmb{B}^{T}\pmb{B} & \frac{1}{p_{2}}\pmb{B}^{T}\pmb{e} \\
\pmb{\mu}_{2} \pmb{e}^{T}\pmb{A}+\frac{1}{p_{2}}\pmb{e}^{T}\pmb{B} & \pmb{\mu}_{2} m_{1}+\frac{1}{p_{2}}m_{2} & \frac{1}{p_{2}}\pmb{e}^{T}\pmb{B} & \frac{1}{p_{2}}m_{2} \\
\pmb{B}^{T}\pmb{B} & \pmb{B}^{T}\pmb{e} & \pmb{e}\tau\pmb{e}^{T}+\pmb{B}^{T}\pmb{B} & \pmb{B}^{T}\pmb{e} \\
\pmb{e}^{T}\pmb{B} & m_{2} & \pmb{e}^{T}\pmb{B} & m_{2}\\
  \end{array}
\right)^{-1} \left(
  \begin{array}{@{}*6r}
\pmb{\mu}_{2} \pmb{A}^{T}\pmb{e} \\
\pmb{\mu}_{2} \\
0 \\
\tau\\
  \end{array}
\right)
$
}
\end{equation}

By finding the equations of the two fuzzy hyperplanes, given a new test sample, the fuzzy distance between the sample and each of the fuzzy hyperplanes are calculated. Definition $1$ defines the fuzzy distance between a data point and a fuzzy hyperplane.

\textbf{Definition 1}: $\Delta=(\delta,\gamma)$ is the fuzzy distance between a data point $\pmb{x}=(x_{1},\cdots,x_{d})^T$ and the fuzzy hyperplane $\w^{T}.\pmb{x}+\bhyper$ with $\w = (\pmb{w},\pmb{c})$ and $\bhyper = (b,d)$, where $\delta$$=\frac{\lvert w_{1}x_{1}+\cdots+w_{d}x_{d}+b\rvert}{\lVert \pmb{w}\rVert}$ and $\gamma$$=\frac{\lvert (w_{1}+c_{1})x_{1}+\cdots+(w_{d}+c_{d})x_{d}\rvert}{\lVert \pmb{w}\rVert}$.

Using the fuzzy distances between a data point and the fuzzy hyperplanes, we define a fuzzy membership function which determines the membership degree of the data point to each class. Assume $\Delta_{1}=(\delta_{1},\gamma_{1})$ and $\Delta_{2}=(\delta_{2},\gamma_{2})$ are fuzzy distances between a data point and the two hyperplanes $H_{1}$ and $H_{2}$, respectively. For an input data $\pmb{x}_{0}$, the degree to which $\pmb{x}_{0}$ belongs to hyperplane $H_{1}$ is defined by the following membership function (knowing the membership degrees for $H_{1}$, the membership degrees for $H_{2}$ are easy to find):
\begin{flalign}\label{eq29}
&\mu_{1}(\pmb{x}_{0})=\begin{cases}
	1-\frac{\delta_{1}+\gamma_{1}}{\delta_{1}+\gamma_{1}+\delta_{2}+\gamma_{2}} & \delta_{1} \geq \gamma_{1}, \delta_{2}\geq \gamma_{2},\\
	1-\frac{\delta_{1}}{\delta_{1}+\delta_{2}+\gamma_{2}} & \delta_{1} < \gamma_{1}, \delta_{2}\geq \gamma_{2},\\
	1-\frac{\delta_{1}+\gamma_{1}}{\delta_{1}+\gamma_{1}+\delta_{2}} & \delta_{1} \geq \gamma_{1}, \delta_{2} < \gamma_{2}\\
	1-\frac{\delta_{1}}{\delta_{1}+\delta_{2}} & \delta_{1} < \gamma_{1}, \delta_{2} < \gamma_{2},\\
\end{cases}&
\end{flalign}

\section{Numerical Results}\label{SecIV}
To evaluate the performance of our proposed algorithm, we investigate its classification accuracy on both artificial and benchmark data sets. 
Accuracies are obtained by the standard 10-fold cross-validation. The focus of our experiments is on comparing the performance of our $M_{2}$ model with SVM and LST-SVM. It should be noted that for the benchmark data sets the parameters of all algorithms, i.e. a single penalty parameter for SVM and P-SVM and two penalty parameters for the rest of algorithms, are tuned as explained in \cite{arun2009least}, while for the toy data set and NDS data sets, all penalty parameters are set to 1. The MATLAB code of the FLST-SVM can be found here (\textit{A link to a public repository will be provided upon acceptance of the paper})

\subsection{Experiment on toy data set}
\begin{figure}[ht!]
	\centering
	\begin{subfigure}[b]{0.45\columnwidth}
		\includegraphics[width=\columnwidth]{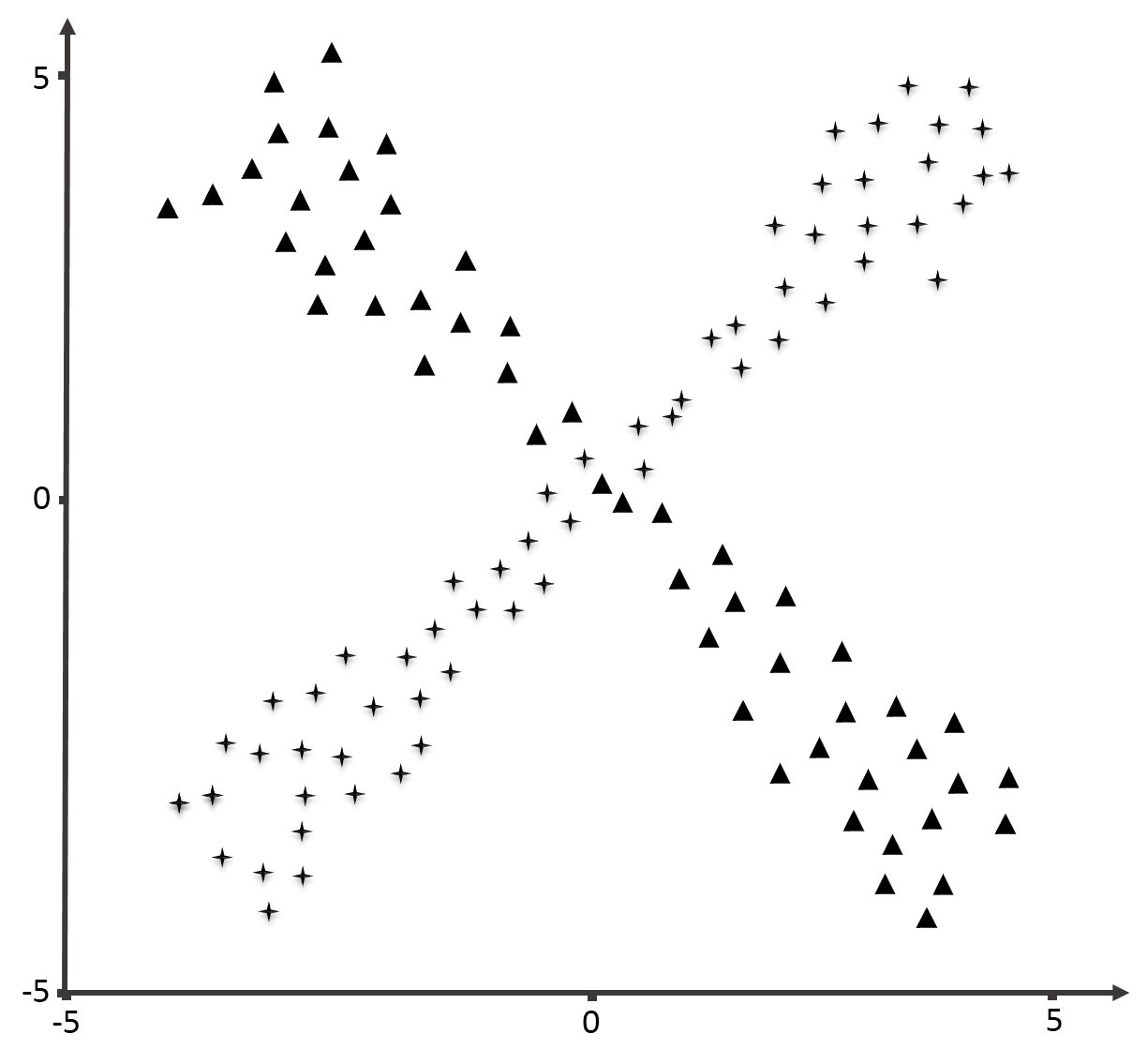}
		\caption{}
		\label{Toydata}
	\end{subfigure}%
	\hspace{0.02\textwidth}%
	\begin{subfigure}[b]{0.45\columnwidth}
		\includegraphics[width=\columnwidth]{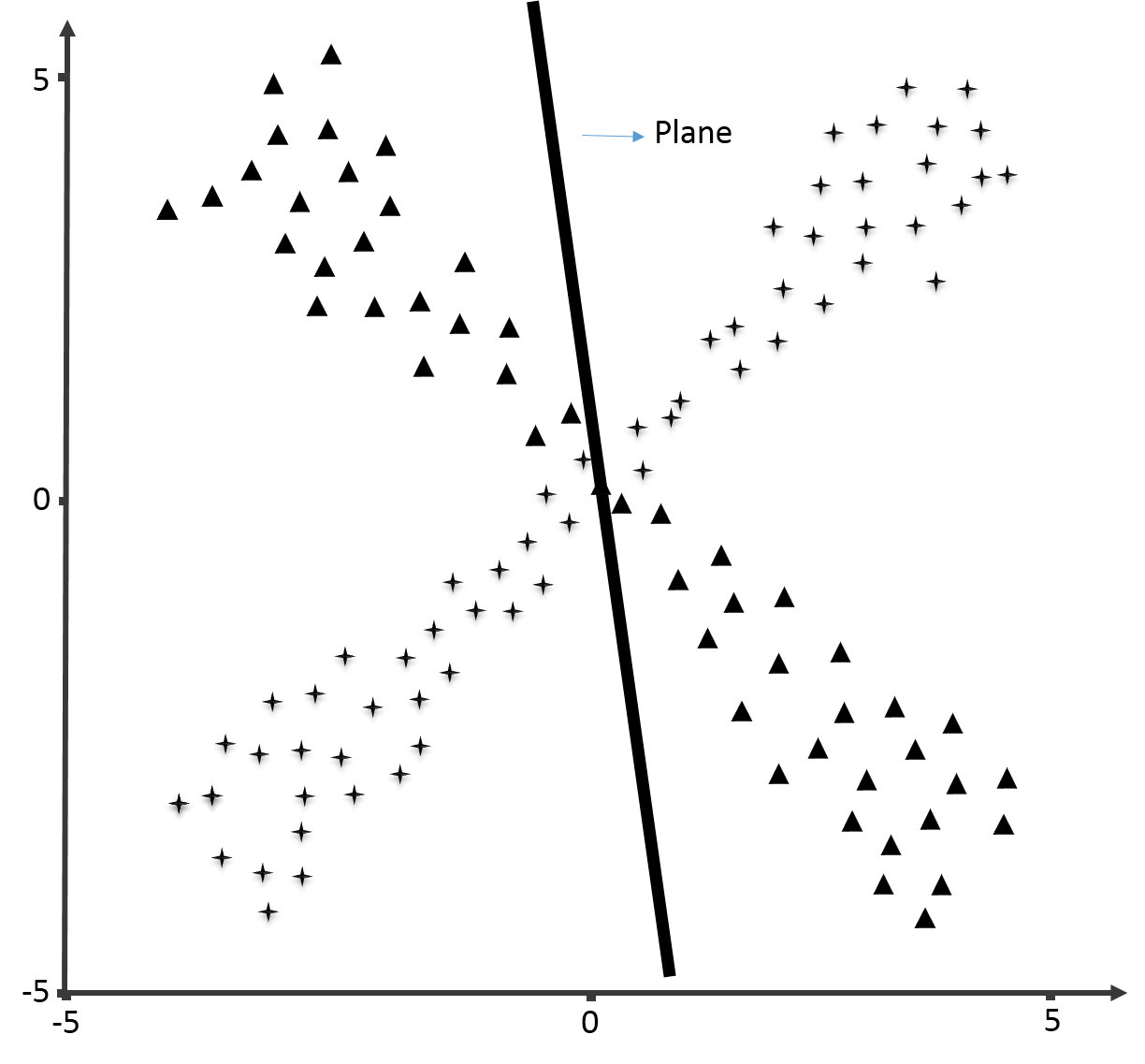}
		\caption{}
		\label{Fig4.a}
	\end{subfigure}%
	\hspace{0.02\textwidth}%
	\begin{subfigure}[b]{0.45\columnwidth}
		\includegraphics[width=\columnwidth]{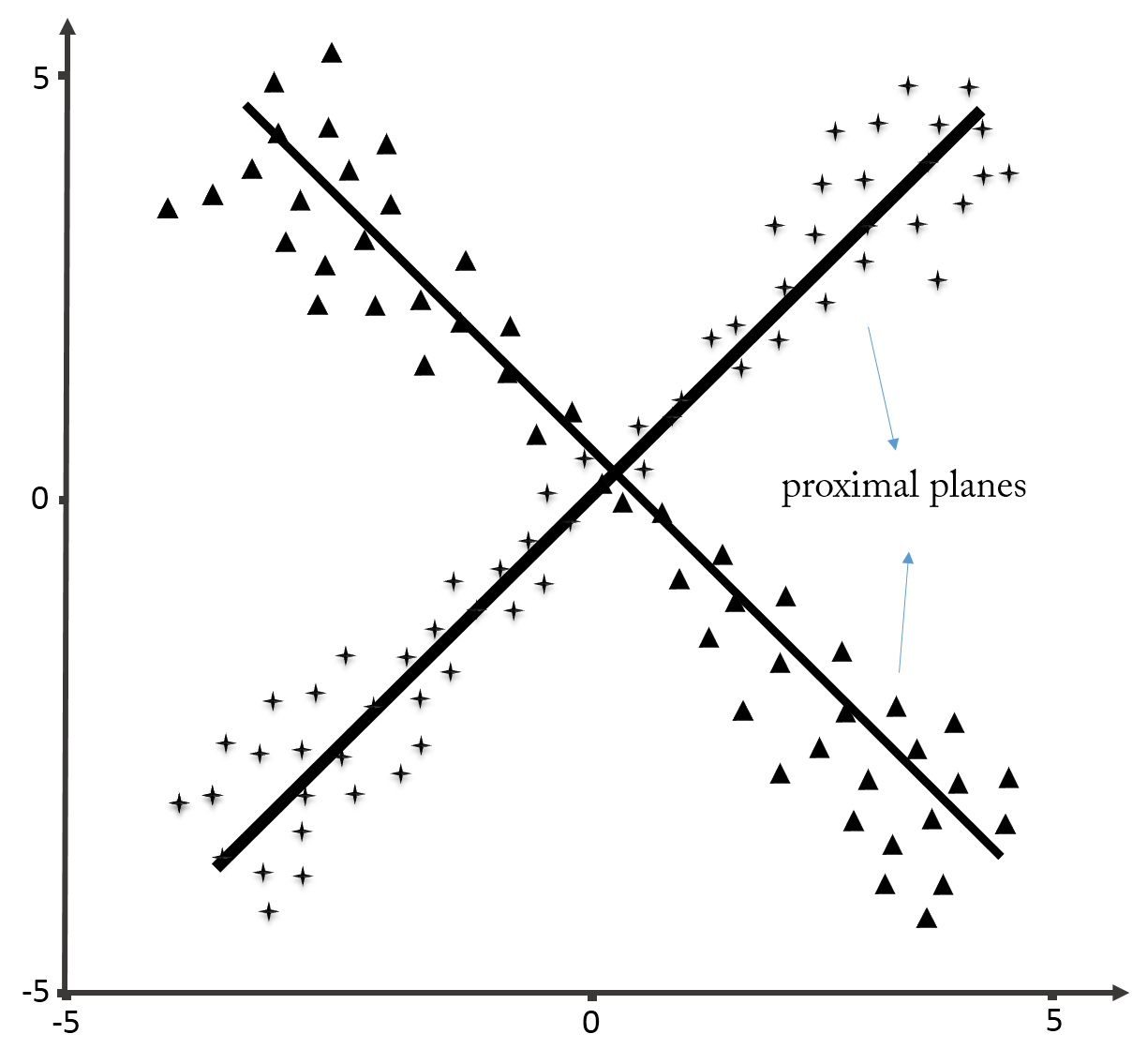}
		\caption{}
		\label{Fig4.b}
	\end{subfigure}
	\hspace{0.02\textwidth}
	\begin{subfigure}[b]{0.45\columnwidth}
		\includegraphics[width=\columnwidth]{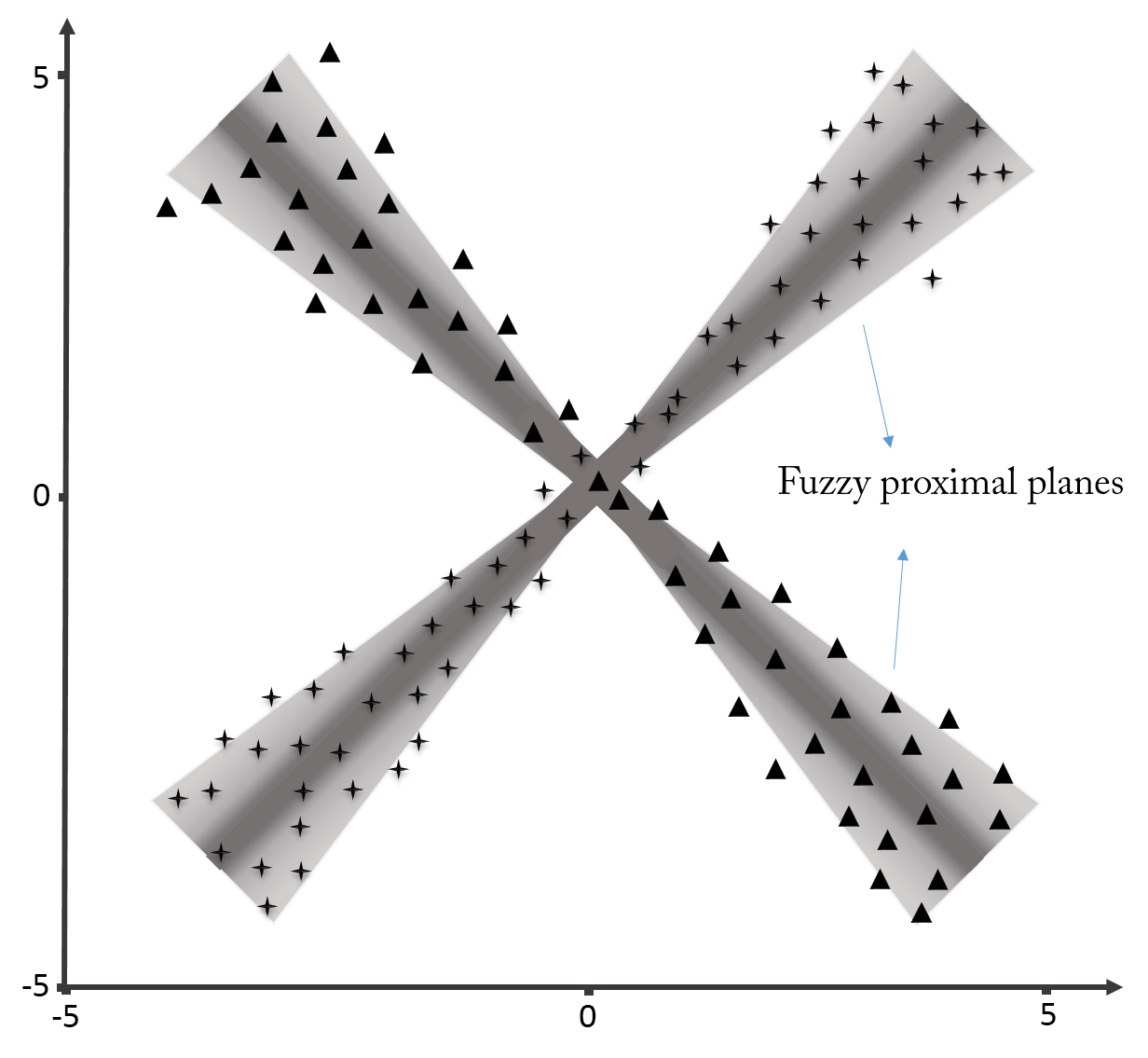}
		\caption{}
		\label{Fig4.c}
	\end{subfigure}
	\caption{The (a) synthetic data set and decision lines obtained by (b) SVM, (c) LST-SVM and (d). FLST-SVM}
	\label{Fig4}
\end{figure}

We adopted the simple two-dimensional XOR data, which is a common example for evaluating the effectiveness of SVM based algorithms, shown in Figure \ref{Toydata}. This hand-made data set consists of 121 records belonging to two classes. Each sample has two features: 1) class and 2) membership degree, which determines to what extend the sample belongs to the class. In figure \ref{Toydata}, stars denote samples of positive class and triangulars are samploes of negative class.

Table \ref{TI} denotes the classification accuracies obtained by applying SVM, LST-SVM and FLST-SVM algorithms to this data set. SVM uses a single hyperplane for classifying data and since space is two-dimensional, this hyperplane is a line as it is shown hypothetically in Figure \ref{Fig4.a}. LST-SVM, on the other hand, has two lines for classification. As mentioned in Section \ref{SecII.3}, these lines are closest to the sample of their corresponding classes and farthest from the samples of the opposite class. Figure \ref{Fig4.b} shows the hypothetical lines of the LST-SVM. Because the samples are not linearly separable, LST-SVM has still large amount of error although it provides higher accuracy when compared to the SVM. FLST-SVM also has two hyperplanes responsible for classifying data with the difference that these two lines are not crisp. Figure \ref{Fig4.c} shows the hypothetical fuzzy lines of FLST-SVM. As the distance from the center of the hyperplane increases, the amount of vagueness increases, and vice versa. To show the fuzzy nature of each line, we have used multiple lines. As shown in Table \ref{TI}, these fuzzy lines discriminate the samples better than SVM and LST-SVM and provide higher accuracy.
\begin{table}[hb!]
	\centering
	\caption{A comparison of classification accuracy. Best performance is \textbf{bolded}}
	\begin{tabular}{c c}
		\hline
		\textbf{Algorithms} & \textbf{Accuracy(\%)}\\
		\hline
		SVM & 53.0\\
		LST-SVM & 65.0\\
		FLST-SVM & \textbf{73.0}\\
		\hline
	\end{tabular}
	\label{TI}
\end{table}
\subsection{Experiments on benchmark data sets}
To illustrate the performance of FLST-SVM on real data sets, we perform experiments on 10 benchmark data sets from the UCI machine learning repository \cite{blake1998uci} with details listed in Table \ref{TIII}. These data sets represent a wide range of different sizes (from 155 to 1473) and different dimensionalities (from 7 to 34). We compared predictive performance of FLST-SVM with SVM, LST-SVM, T-SVM, P-SVM, and GEP-SVM. In this experiment, we employ two different criteria to assign fuzzy membership values to each sample in the data sets. Assume a data set with $n$ samples, $\{\pmb{x}_i\}_{i=1}^n$, and two classes $+$ and $-$. Let $\pmb{A}$ and $\pmb{B}$ be matrices of samples belong to class $+$ and class $-$, respectively. The fuzzy membership $\mu_i$ for each sample of class $+$ is given by:
\begin{align}
\mu_i=\frac{\pmb{x}_i - c_{+}^{Mean}}{r_{+}^{Max}+\epsilon}
\end{align}
where $r^{+}_{Max}=\max |\pmb{x}_i-c_{+}^{Mean}|, \; \pmb{x}_{i}\in \pmb{A}$, $c_{+}^{Mean}= \frac{1}{\mid \pmb{A} \mid} \sum_{\pmb{x}_i\in \pmb{A}}\pmb{x}_i$, and $\epsilon$ is a small constant value to avoid dividing by zero. Membership degrees of the negative class are obtained in a similar way. Another way of associating fuzzy membership degrees to each sample is to run LST-SVM and compute the distance of each sample from its hyperplane as the fuzzy membership.
\begin{table*}[ht!]
	\centering
	\caption{10-fold cross-validation \textit{mean accuracy $\pm$ standard deviation} of different algorithms on different UCI data sets. Best performances are shown in bold face.}
	\label{TIII}
	\begin{adjustbox}{width=1\textwidth}
		\small
		\begin{tabular}{l*{6}{c}r}
			\hline
			Dataset              & FLST-SVM ($\pmb{1}$) & LST-SVM ($\pmb{2.95}$) & SVM ($\pmb{5.35}$) & T-SVM ($\pmb{3.5}$) & P-SVM ($\pmb{4.1}$)  & GEP-SVM ($\pmb{4.0}$) \\
			\hline
			Pima Indians Diabetes (768 $\times$ 8) & \textbf{79.63 $\pm$ 3.5} & 77.3 $\pm$ 2.8 & 72.6 $\pm$ 1.3 & 73.4 $\pm$ 6.4 & 75.8 $\pm$ 3.8 & 74.2 $\pm$ 3.9  \\
			Heart-Statlog (270 $\times$ 13)       & \textbf{88.7 $\pm$ 3.2} & 83.4 $\pm$ 4.2 & 81.2 $\pm$ 2.9 & 83.2 $\pm$ 3.1 & 83.6 $\pm$ 5.4 & 85.1 $\pm$ 6.5 \\
			Australian (690 $\times$ 14)          & \textbf{92.9 $\pm$ 4.5} & 82.7 $\pm$ 3.8 & 82.2 $\pm$ 2.3 & 83.5 $\pm$ 3.6 & 83.1 $\pm$ 4.1 & 78.7 $\pm$ 4.3  \\
			Heart-c (303 $\times$ 14)     & \textbf{87.6 $\pm$ 2.9} & 82.9 $\pm$ 5.6 & 79.5 $\pm$ 2.8 & 82.8 $\pm$ 5.3 &  62.5 $\pm$ 6.1 & 81.7 $\pm$ 6.3  \\
			Bupa Liver Disorder (345 $\times$ 7)     & \textbf{79.6 $\pm$ 5.4} & 68.7 $\pm$ 4.2 & 69.4 $\pm$ 1.8 & 69.4 $\pm$ 4.1&  68.8 $\pm$ 4.3 & 66.2 $\pm$ 4.4  \\
			Hepatitis (155 $\times$ 19)     & \textbf{91.3 $\pm$ 4.2} & 85.4 $\pm$ 8.2 & 80.7 $\pm$ 3.8 & 84.5 $\pm$ 5.8 &  83.9 $\pm$ 5.2 & 84.2 $\pm$ 9.2  \\
			Wisconsin Breast Cancer (198 $\times$ 34)    & \textbf{98.2 $\pm$ 1.6} & 83.7 $\pm$ 5.6 & 81.2 $\pm$ 2.4 & 83.2 $\pm$ 6.3 &  83.3 $\pm$ 4.2 & 79.8 $\pm$ 8.2  \\
			CMC (1473 $\times$ 9)    & \textbf{75.4 $\pm$ 4.8} & 67.8 $\pm$ 3.1 & 65.6 $\pm$ 2.3 & 67.8 $\pm$ 3.2 &  67.1 $\pm$ 2.9 & 68.1 $\pm$ 3.9 \\
			Votes (435 $\times$ 16)    & \textbf{98.7 $\pm$ 4.6} & 94.6 $\pm$ 2.8 & 88.4 $\pm$ 2.5 & 93.1 $\pm$ 3.4 & 92.8 $\pm$ 3.5 & 93.0 $\pm$ 3.4  \\
			Sonar (208 $\times$ 60)    & \textbf{95.1 $\pm$ 3.7} & 79.2 $\pm$ 5.9 & 76.9 $\pm$ 2.7 & 78.7 $\pm$ 5.1 &  78.9 $\pm$ 4.9 & 79.8 $\pm$ 7.6 \\
			\hline
		\end{tabular}
	\end{adjustbox}
	
\end{table*}

Table \ref{TIII} compares the results of different algorithms. Results have been reported as the mean of the accuracies obtained in the 10-fold cross-validation plus/minus the standard deviation. As shown in the table, FLST-SVM outperforms all the other versions in all data sets.

However, it is not conclusive to just compare the accuracies; we need to verify whether or not the improvements made by the FLST-SVM are statistically significant. To do so, we first investigate if the differences between the accuracies of the algorithms are statistically meaningful or not. Just after that, we check whether or not our FLST-SVM is significantly better than other algorithms in predicting the class labels.

To find the answer to the first question above, i.e. are the difference shown in Table \ref{TIII} statistically meaningful, we used the Friedman test which is a non-parametric counterpart of ANOVA. The Friedman test ranks the algorithms for each data set separately, based on their performance so that the best performing algorithm getting rank 1 and position 1, the second best ranked 2 and position 2, and so on. In case of ties the average of the positions is adopted as the rank of the algorithms. As an example, in CMC data set, LST-SVM and T-SVM are both $3^{rd}$ rank algorithm according to the accuracies. However, their positions are $3$ and $4$ (after FLST-SVM in the first position and GEP-SVM in the second position), so both will be ranked $3.5$ and the next algorithm will be $5^{th}$. Then the average ranks of each classifier over all data sets is computed, and the Friedman statistic is calculated (equation \ref{eq24}). Assuming $C$ and $D$ are the total number of classifiers to be compared (in our case $6$), and the total number of data sets (in our case $10$), respectively, the Friedman statistic follows a $\chi^{2}_{F}$ distribution with $C-1$ degrees of freedom, when $D$ and $C$ are large enough (as a rule of thumb, $D >= 10$ and $C >= 5$), which are in our case (for more details check \cite{demvsar2006statistical}). Finally, the critical value of the $\chi^{2}_{F}$ distribution is compared with the statistic itself. Assuming that the null hypothesis is that there is no significant difference among different algorithms in their predictive performance, it will be rejected if the statistic is higher than the critical value.

The formula for Friedman statistic is 
\begin{equation}\label{eq24}
	\chi_{F}^{2} = \frac{12D}{C(C+1)}\left[\sum_{j}^{}R_{j}^{2}-\frac{C(C+1)^{2}}{4}\right]
\end{equation}
where $R_{j}$ is the average rank of the $j^{th}$ classifier. The average rank of each classifier is shown in bold face inside the parenthesis in the first row of Table \ref{TIII} and the Friedman statistic computed from the table is $28.24$ and the critical value of $\chi^{2}_{F}$ with $5$ degrees of freedom and with significance level of $0.05$ is $11.070$. Since the critical value is smaller than the Friedman statistic of our results, we reject the null-hypothesis, meaning that the algorithms are statistically different.

By rejecting the null-hypothesis of the Friedman test, we can proceed with a post-hoc test to analyze whether or not our FLST-SVM is significantly better than each of the other algorithms. For this goal, we use Nemenyi test \cite{nemenyi1963distribution}. The performance of two classifiers is significantly different if their ranks differ by at least the critical difference of
\begin{equation}
	\mbox{CD} = q_{\alpha}\sqrt{\frac{C(C+1)}{6D}}
\end{equation}
where $q_{\alpha}$ is the critical value for the two-tailed Nemenyi test \cite{demvsar2006statistical}. The critical value for six classifiers and with significance level of $0.05$ is $2.850$ and, therefore, we have $\mbox{CD} = 2.850\sqrt{\frac{6\times 7}{6\times 10}} = 2.38$. Using this critical difference, we conclude that:
\begin{itemize}
	\setlength\itemsep{0em}
	\item The difference between FLST-SVM and LST-SVM is not statistically significant, since $2.95 - 1 \ngeq 2.38$
	\item The difference between FLST-SVM and Lib-SVM is statistically significant, since $5.35 - 1 \geq 2.38$
	\item The difference between FLST-SVM and T-SVM is statistically significant, since $3.5 - 1 \geq 2.38$
	\item The difference between FLST-SVM and P-SVM is statistically significant, since $4.1 - 1 \geq 2.38$
	\item The difference between FLST-SVM and GEP-SVM is statistically significant, since $4 - 1 \geq 2.38$
\end{itemize}
We argue that the reason why the difference between FLST-SVM and LST-SVM is not significant might be due to the critical value of the Nemenyi test which is adjusted for making $C(C-1)/2$ comparisons while we only make $C-1$ comparisons. However, we found no other test that suits our problem better than Nemenyi test.

\subsection{Experiments with large data sets}

We also conducted experiments on larger data sets, generated using NDC data generator \cite{ndc}. Table \ref{TV} describes characteristics of the generated NDC data sets.
\begin{table}[h!]
	\centering
	\caption{Details of generated NDC data sets}
	\label{TV}
	\begin{tabular}{l*{3}{c}r}
		Dataset              & \#Features& \#Training samples & \#Test samples  \\
		\hline 
		NDC-1k  & 32 & 1000 & 100 \\
		NDC-5k  & 32 & 5000 & 500 \\
		NDC-10k & 32 & 10000 & 1000  \\
		NDC-1m  & 32 & 1000000 & 100000 \\		
	\end{tabular}
\end{table}
Table \ref{TVI} represents comparison of accuracies of FLST-SVM with other versions of SVM. Results of LST-SVM, T-SVM, P-SVM and GEP-SVM are taken from \cite{arun2009least} and the results of SVM are computed using WEKA \cite{Hall2009WDM}. The results demonstrate that FLST-SVM outperforms LST-SVM and other versions of SVM in terms of generalization. We do not run the significance test for this experiment due to the small number of data sets ($D < 10$), and also because of the absence of the performance of SVM and T-SVM for NDC-1m. These will make the results of the significance test unreliable. Although, we expect to have the same results as for the significance test in the previous experiment since FLST-SVM has the best performance for all data set sizes. Finally, from the run time point of view, both FLST-SVM and LST-SVM have the same running time and perform several times faster than TSVM and SVM on all data sets. This is because FLST-SVM and LST-SVM do not use any optimizer in contrast with SVM, TSVM and GEP-SVM.
\begin{table*}[ht!]
	\centering
	\caption{Accuracies of the algorithms on the generated NDC data sets.}
	\label{TVI}
	\begin{adjustbox}{width=1\textwidth}
		\small
		\begin{tabular}{l*{6}{c}r}
			\hline
			Dataset              & FLST-SVM  & LST-SVM  & SVM  & T-SVM  & P-SVM   & GEP-SVM \\
			\hline
			NDC-1k     & \textbf{89.94} & 88.00 & 66.66 & 83.00 &  80.00 & 69.00  \\
			NDC-5k     & \textbf{85.39} & 80 & 64.82 & 79.80 &  80 & 74.4  \\
			NDC-10k     & \textbf{87.59} & 87.00 & 71.56 & 87.3 &  87.00 & 83.9  \\
			NDC-1m     & \textbf{86.75} & 86.24 & $^*$ & $^*$ &  86.24 & 84.12 \\
			\hline
			\\
		\end{tabular}
	\end{adjustbox}
	
	$^*$ We stopped experiments as computing time was very high.
\end{table*}
\section{Conclusion}\label{SecV}
In this paper, we enriched the LST-SVM classifier by incorporating the theory of fuzzy sets. We proposed two novel models for fuzzy LST-SVM. In the first model, $M_{1}$, a fuzzy membership degree is assigned to each sample and the hyperplanes are optimized based on the fuzzy importance degrees. In the second model, $M_{2}$, all parameters to be learned in LST-SVM are considered fuzzy parameters and two fuzzy hyperplanes are learned. We conducted a series of experiments to compare our classifier with SVM and LST-SVM. The results demonstrate that FLST-SVM significantly improves the classification accuracies.

FLST-SVM could be applied into different applications such as disease detection, image analysis, spam filtering, weather forecasting, and intrusion detection. In addition to binary classification, it can applied to multi-class classification problems. 

Finally, it should be noted that the focus of our work in this paper is on linear LST-SVM. We proposed two linear models for FLST-SVM and compare the performance of one of them with different linear versions of SVM. Deriving the non-linear FLST-SVM model is also straightforward and could be accomplished in the same way, but requires more careful attention. This will be a possible future direction of the domain.

%
\bibliography{Reference}
\end{document}